\documentclass[11pt]{article}

\usepackage[final]{acl}
\usepackage{times}
\usepackage{latexsym}
\usepackage[T1]{fontenc}
\usepackage[utf8]{inputenc}
\usepackage{makecell}
\usepackage{microtype}
\usepackage{pifont}
\usepackage{inconsolata}

\usepackage{graphicx}

\usepackage{algorithm}
\usepackage{algorithmic}
\usepackage{amsmath}
\usepackage{color, xspace}
\usepackage{enumitem}
\usepackage{verbatim}
\usepackage{amssymb}
\usepackage{booktabs}
\usepackage{multirow}
\usepackage{xcolor}
\usepackage{tcolorbox}

\usepackage[normalem]{ulem}
\useunder{\uline}{\ul}{}

\tcbuselibrary{skins}
\definecolor{sectionblue}{RGB}{0, 82, 155}
\definecolor{keywordpurple}{RGB}{128, 0, 128}
\definecolor{lightgray}{RGB}{245, 245, 245}
\definecolor{darkgray}{RGB}{80, 80, 80}

\newcommand{\mname}{Tandem\xspace}
\newcommand*\samethanks[1][\value{footnote}]{\footnotemark[#1]}

\DeclareMathOperator*{\argmax}{arg\,max}
\usepackage{pgf}

\colorlet{GainText}{green!50!black}
\colorlet{LossText}{red!60!black}

\newcommand{\accpct}[2]{%
  \pgfmathparse{((#1-#2)/#2)*100}%
  \ifdim \pgfmathresult pt > 0pt
    \pgfmathparse{ceil(\pgfmathresult)}%
    \pgfmathtruncatemacro{\diff}{\pgfmathresult}%
    {\footnotesize\textcolor{GainText}{(+\diff\%)}}%
  \else\ifdim \pgfmathresult pt < 0pt
    \pgfmathparse{floor(\pgfmathresult)}%
    \pgfmathtruncatemacro{\diff}{\pgfmathresult}%
    \pgfmathtruncatemacro{\absdiff}{abs(\diff)}%
    {\footnotesize\textcolor{LossText}{(-\absdiff\%)}}%
  \else
  \fi\fi
}

\newcommand{\downpct}[2]{%
  \pgfmathparse{((#1-#2)/#2)*100}%
  \ifdim \pgfmathresult pt < 0pt
    \pgfmathparse{floor(\pgfmathresult)}%
    \pgfmathtruncatemacro{\diff}{\pgfmathresult}%
    \pgfmathtruncatemacro{\absdiff}{abs(\diff)}%
    {\footnotesize\textcolor{GainText}{(-\absdiff\%)}}%
  \else\ifdim \pgfmathresult pt > 0pt
    \pgfmathparse{ceil(\pgfmathresult)}%
    \pgfmathtruncatemacro{\diff}{\pgfmathresult}%
    {\footnotesize\textcolor{LossText}{(+\diff\%)}}%
  \else
  \fi\fi
}

\title{Tandem: Riding Together with Large and Small Language Models \\ for Efficient Reasoning}
\author{
  \textbf{Zichuan Fu\textsuperscript{1}\thanks{Work was conducted during the internship at Tencent.}},
  \textbf{Xian Wu\textsuperscript{2}\thanks{Corresponding author.}},
  \textbf{Guojing Li\textsuperscript{1,3}},
  \textbf{Yejing Wang\textsuperscript{1}},
  \textbf{Yijun Chen\textsuperscript{1}},\\
  \textbf{Zihao Zhao\textsuperscript{1}},
  \textbf{Yixuan Luo\textsuperscript{1}},
  \textbf{Hanyu Yan\textsuperscript{1}},
  \textbf{Yefeng Zheng\textsuperscript{4}},
  \textbf{Xiangyu Zhao\textsuperscript{1}\samethanks}
\\
\\
  \textsuperscript{1} City University of Hong Kong
  \textsuperscript{2} Tencent Jarvis Lab  \\
  \textsuperscript{3} Renmin University of China 
  \textsuperscript{4} Westlake University
\\
  \small{
  \texttt{
    \href{mailto:zc.fu@my.cityu.edu.hk}{zc.fu@my.cityu.edu.hk},
    \href{mailto:kevinxwu@tencent.com}{kevinxwu@tencent.com},
    \href{mailto:xianzhao@cityu.edu.hk}{xianzhao@cityu.edu.hk}
  }}
}

\begin{document}
\maketitle
\begin{abstract}
% Recent advancements in large language models have led to the widespread adoption of thinking paradigms, where models perform explicit reasoning before generating final answers. While this approach improves the response quality, it significantly increases computational costs due to extended output lengths. 
% In this paper, we propose \mname, a large--small language model collaboration framework for low-cost yet high-quality reasoning.
% Specifically, the large language model (LLM) provides guidance by generating four types of critical reasoning insights during the thinking phase. Based on the specific guidance, the small language model (SLM) continues the reasoning process and produces the final answer.
% To reduce redundant computation, \mname employs a cost-aware judgment mechanism that uses perplexity and entropy as confidence indicators to adaptively determine whether sufficient insights have been accumulated, allowing the LLM to terminate its reasoning early.
% Experiments on public datasets demonstrate that our approach saves approximately half of the computational cost while achieving superior performance compared to the LLM alone.
% The code is available at: %\href{https://anonymous.4open.science/r/Ensemble-Hub-0FD8}{anonymous repository}.
% \url{https://anonymous.4open.science/r/Ensemble-Hub-0FD8}

Recent advancements in large language models (LLMs) have catalyzed the rise of reasoning-intensive inference paradigms, where models perform explicit step-by-step reasoning before generating final answers. 
While such approaches improve answer quality and interpretability, they incur substantial computational overhead due to the prolonged generation sequences.
%While this approach enhances response quality, it significantly increases computational costs due to the extended length of generated sequences.
In this paper, we propose \mname, a novel collaborative framework that synergizes large and small language models (LLMs and SLMs) to achieve high-quality reasoning with significantly reduced computational cost.
% a collaborative framework between large and small language models designed to achieve high-quality reasoning with low computational overhead. 
Specifically, the LLM serves as a strategic coordinator,
efficiently generating a compact set of critical reasoning insights. These insights are then used to guide a smaller, more efficient SLM in executing the full reasoning process and delivering the final response. 
% the LLM acts as a guide by efficiently generating four types of critical reasoning insights during the planning phase. Guided by these insights, a small language model (SLM) executes the subsequent reasoning process and produces the final answer.
% To maximize computational efficiency, % 这个target 不对吧
To balance efficiency and reliability, \mname introduces a cost-aware termination mechanism that adaptively determines when sufficient reasoning guidance has been accumulated, enabling early stopping of the LLM’s generation.    
% This mechanism leverages confidence indicators such as perplexity and entropy to dynamically assess reasoning completeness. % 技术细节，可以省略
% To balance the computational efficiency and response reliability, \mname incorporates a cost-aware judgment mechanism. 
% By utilizing perplexity and entropy as confidence indicators, this mechanism adaptively determines when sufficient insights have been accumulated, allowing the LLM to terminate its generation early. 
Experiments on mathematical reasoning and code generation benchmarks demonstrate that \mname reduces computational costs by approximately 40\% compared to standalone LLM reasoning, while achieving superior or competitive performance. Furthermore, the sufficiency classifier trained on one domain transfers effectively to others without retraining.
The code is available at: \href{https://github.com/Applied-Machine-Learning-Lab/ACL2026_Tandem}{https://github.com/Applied-Machine-Learning-Lab/ACL2026\_Tandem}.
% \url{https://anonymous.4open.science/r/Ensemble-Hub-0FD8}

\end{abstract}

\section{Introduction}
\label{introduction}

\begin{figure}[t]
    \centering
    \includegraphics[width=\linewidth]{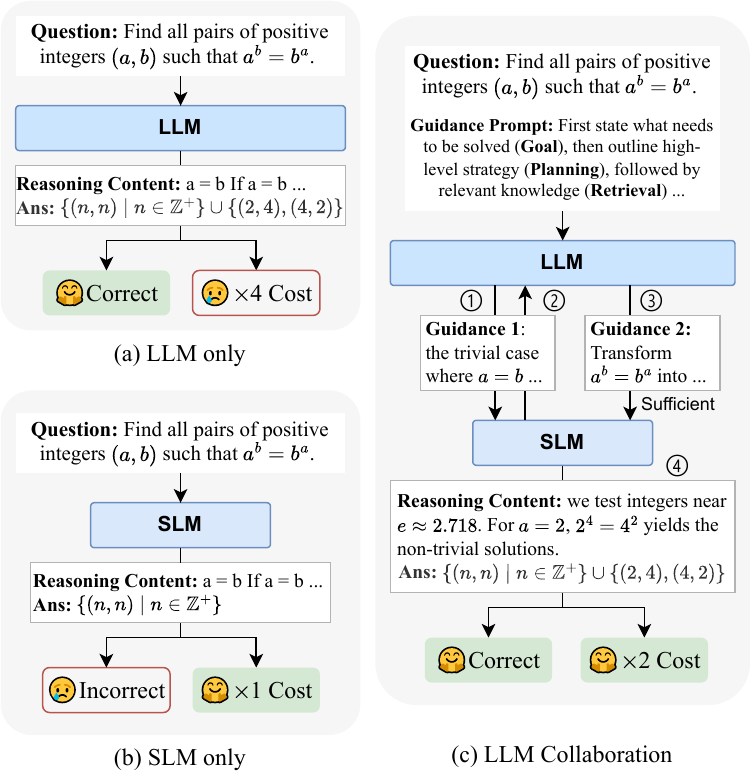}
    \caption{Comparison of reasoning-inference strategies:
        (a) LLM-only. 
        (b) SLM-only. 
        (c) LLM-SLM collaboration, where the LLM provides guidance and the SLM completes the reasoning.}
    \label{fig:act-e}
\end{figure}

% First paragraph
% Background of LLM Development and the Demand for Ensemble
Recent advances in large language models (LLMs) have shifted their role from simple text generators to sophisticated reasoning systems capable of solving complex problems~\cite{reasoning-survey,llm-enhanced-recsys,rethink-llm-seqrec}. A key development in this shift is the emergence of a thinking paradigm, which extends earlier chain-of-thought~\cite{cot} prompting. In this paradigm, exemplified by models such as DeepSeek-R1~\cite{deepseekr1}, the internal reasoning process is explicitly externalized and structurally decoupled from final answer generation. This design substantially improves interpretability and faithfulness, and has led to state-of-the-art performance on challenging mathematical and scientific reasoning tasks.

% Second paragraph
% Limitations of Traditional Ensemble Methods
However, these reasoning improvements come at a substantial computational cost. 
Thinking models routinely generate reasoning chains spanning thousands of tokens, typically 5--10 times longer than conventional LLM outputs~\cite{overthinking}. 
This extended reasoning leads to a dramatic increase in both inference latency and operational expenses. Such overhead poses a major obstacle to real-world deployment, where applications often require real-time responses or operate under strict budget constraints~\cite{hybrid,Reasoning}.
This raises a central question: \textit{how can we preserve the benefits of explicit thinking while making it computationally efficient enough for practical deployment?}

% Third paragraph - Limitations of existing approaches
Since the inference cost of thinking LLMs is dominated by the length of generated reasoning~\cite{Budget}, prior work has attempted to reduce this overhead through reinforcement fine-tuning (RFT), encouraging the model to solve problems with shorter reasoning traces~\cite{thinkless}.
However, such approaches still have limitations: (1) they require continual training of the LLM, and may degrade the model's general capabilities~\cite{rl-llm,navigate-unknown}, and (2) they are inapplicable to closed-source models where only API access is available. 
These constraints motivate us to explore alternatives without modifying the LLM.

% Fourth paragraph - Large-small LLM collaboration
To address this challenge, we propose a collaborative paradigm between large and small language models (LLMs and SLMs) as a promising path toward cost-effective reasoning. As shown in Figure~\ref{fig:act-e} (a) and (b), relying solely on an LLM yields correct answers but incurs high computational cost, whereas an SLM alone is efficient yet prone to errors.
Our key insight is to combine the strengths of both: the LLM acts as a strategic mentor, providing high-level reasoning guidance, while the SLM serves as an agile intern, efficiently executing the detailed reasoning steps and generating the final response. This division of labor enables accurate reasoning at a fraction of the computational cost as in Figure~\ref{fig:act-e} (c).

% To address this challenge, we explore the collaboration between large and small LLMs as a promising direction.
% As illustrated in Figure~\ref{fig:act-e}, using an LLM alone can generate correct answer but incurs substantial computational cost, while using an SLM alone is cost-efficient but often produces incorrect answers.
% By pairing a large, high-capacity model with a smaller, more efficient one, we can leverage the complementary strengths of both: the LLM provides high-level guidance, and the SLM completes the reasoning process to generate the final answer.
% This collaboration effectively reduces the computational burden while still maintaining the reasoning quality of the larger model.

% Fifth paragraph - Our solution
Based on this insight, we present \mname, a novel collaborative reasoning framework that formalizes this dynamic as a mentor-intern architecture~\cite{mentor}. In \mname, the mentor LLM generates lightweight, early-stage thinking insights, which are then used to guide the intern SLM in constructing the full response. 
% a collaborative reasoning framework that optimizes the synergy between large and small language models (LLMs and SLMs) for cost-aware reasoning. 
% This architecture organizes the collaboration into a \textbf{mentor-intern relationship}, where a high-capacity ``mentor'' LLM provides strategic reasoning insights while a lightweight ``intern'' SLM executes the detailed response completion~\cite{mentor}. 
Our design is inherently aligned with the modular nature of human cognition as described in the ACT-R cognitive architecture~\cite{act-r}. Specifically, we decompose the LLM's reasoning into four \textbf{thinking insights}---Goal, Planning, Retrieval, and Action---which correspond to the essential stages of problem-solving: understanding objectives, outlining strategies, gathering knowledge, and executing logic~\cite{qa-llm-agent,adaptive-kgqa,radio-rag,align-grag,mill-qexpand}. 
By transferring only high-level insights instead of full reasoning chains, \mname retains the cognitive depth of the LLM while significantly reducing computational overhead by delegating the remaining reasoning tasks to the SLM.
% By transferring only these high-level insights rather than full-length reasoning chains, \mname retains the cognitive depth of the LLM while significantly reducing computational overhead by offloading the remaining reasoning workload to the SLM.
To further enhance efficiency, \mname introduces a \textbf{cost-aware judgment mechanism} that evaluates whether the current insights are adequate for the SLM to complete the task, allowing for adaptive control over the amount of guidance provided by the LLM.
% To adaptively control the amount of guidance provided by the LLM, \mname introduces a \textbf{cost-aware judgment mechanism} that assesses whether the current insights are sufficient for the SLM to complete the task. 
% The LLM generates guidance in three successive stages with increasing thinking effort levels, and once the guidance is judged sufficient, further LLM reasoning is terminated and the SLM completes the response. 
% This strategy allows the framework to flexibly adjust LLM reasoning length while delegating the remaining computation to the SLM, thereby substantially reducing overall inference cost.
\begin{figure*}[t]
    \centering
    \includegraphics[width=\linewidth]{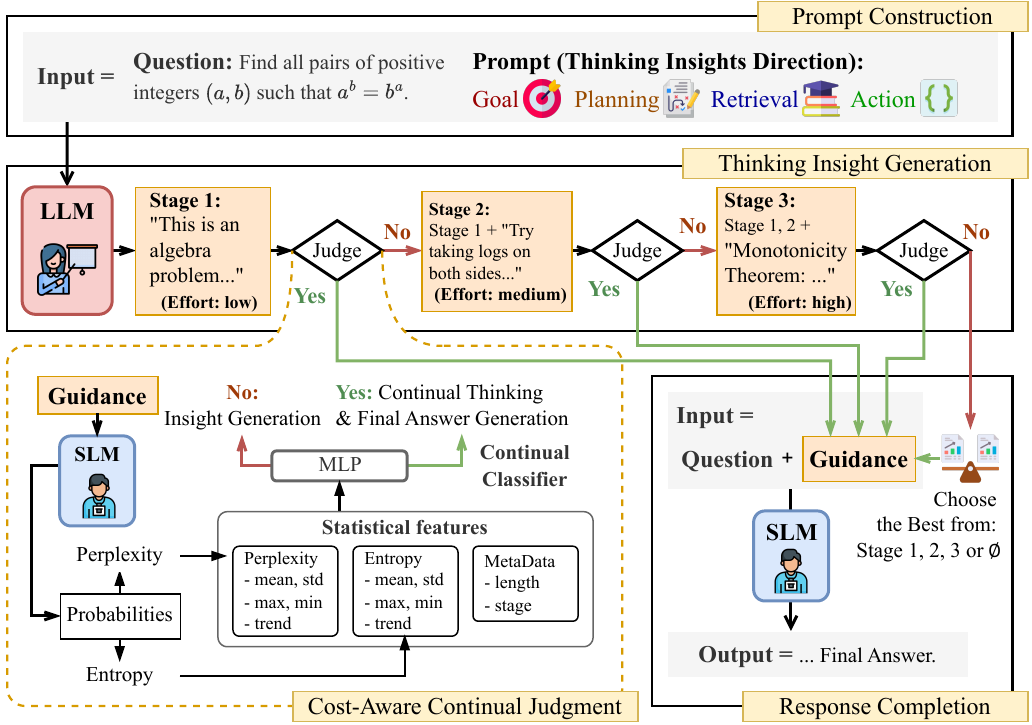}
    \caption{Workflow of the \mname framework. The LLM generates pre-defined thinking insights (Goal, Planning, Retrieval, Action) across three thinking effort levels, while the SLM performs cost-aware continual judgment based on perplexity and entropy to decide when reasoning is sufficient and then completes the final response. %his design achieves efficient and cost-effective reasoning collaboration between large and small models.
    }
    \label{fig:thinking-frame}
\end{figure*}

% Sixth paragraph
% Main Contributions
Our main contributions can be summarized in the following three aspects:
\begin{itemize}[leftmargin=*, itemsep=0pt]
    \item We propose a novel collaborative reasoning paradigm inspired by the mentor–intern relationship, enabling efficient collaboration between large and small models for reasoning tasks.
    \item We propose the \mname framework, which extracts key thinking insights from the LLM to guide the SLM, with a cost-aware judgment mechanism that enables the SLM to adaptively control the insight generation process.
    \item Experiments on mathematical reasoning and code generation benchmarks show that a 32B--7B language model collaboration achieves 2.56\% higher accuracy than the 32B LLM alone on MATH while requiring only 59\% of its computational cost, and the sufficiency classifier transfers across domains without retraining.

\end{itemize}

\section{Method}
\label{sec:method}

\subsection{Framework Overview}
\label{ssec:framework}

As shown in Figure~\ref{fig:thinking-frame}, \mname establishes a collaborative reasoning process between a large language model (LLM) and a small language model (SLM). 
Given a question $Q$, the LLM generates reasoning insights through three\footnote{In this paper, we illustrate \mname through three stages, which can also be extended to include additional stages.} sequential stages of increasing cognitive effort, with later stages providing progressively richer content.  
Insights accumulated up to stage $t$ are aggregated into an insight set $\mathcal{I}^t$. After each stage, the SLM computes uncertainty-based features from the pair $(Q,\mathcal{I}^t)$, and a dedicated classifier $\mathcal{C}$ determines whether the current guidance suffices for response completion. If the classifier predicts sufficiency, the LLM terminates generation and the SLM produces the final answer $A$; otherwise, the LLM proceeds to the subsequent stage for more in-depth insights. 

\begin{figure*}[t]
    \centering
    \includegraphics[width=\linewidth]{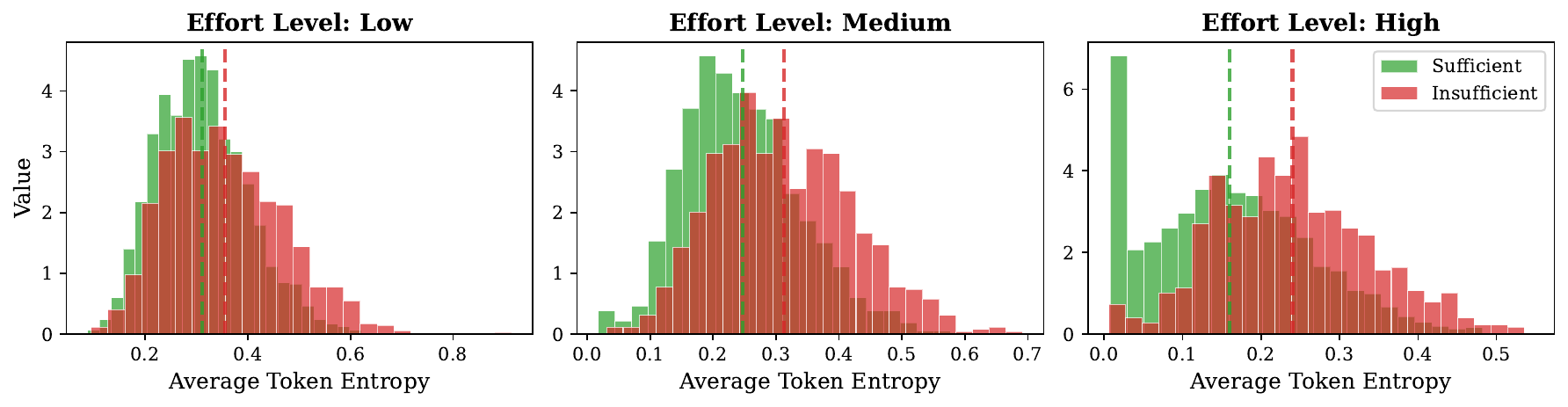}
    \caption{Entropy distribution comparison between sufficient (correct answer) and insufficient (incorrect answer) guidance across three effort levels. Higher effort levels show clearer separation.}
    \label{fig:entropy_distribution}
\end{figure*}

\subsection{Thinking Insight Generation}
\label{ssec:insight}

A straightforward approach to constructing the collaborative thinking-inference paradigm involves directly feeding the LLM's reasoning chains to the SLM to generate responses. However, LLMs typically produce lengthy reasoning traces that include trial-and-error explorations and verbose justifications. Delivering such unprocessed chains to the SLM incurs significant computational overhead and may exceed its comprehension capacity. To mitigate this burden, we extract only the core reasoning components essential for guiding SLMs.

Inspired by the modular nature of human cognition, where cognitive processes are decomposed into distinct functional modules~\cite{act-r}, and by established pipelines for question-answering processes in LLM-based agents~\cite{qa-llm-agent}, we identify and design four types of thinking insights that capture the essential components of effective problem solving:
\begin{enumerate}[leftmargin=*, itemsep=0pt]
    \item \textbf{Goal}: Defines the ultimate objective or question to be solved, clarifying what the model aims to achieve through reasoning.
    \item \textbf{Planning}: Outlines the high-level reasoning strategy, including decomposition of subproblems and selection of solution paths.
    \item \textbf{Retrieval}: Involves recalling or gathering relevant knowledge, facts, or contextual information necessary for problem solving.
    \item \textbf{Action}: Executes concrete reasoning steps, calculations, or logical operations.
\end{enumerate}
These components form the basis of our structured prompt $p$ (detailed in Appendix~\ref{app:prompt_schema}) that instructs the LLM to generate insights sequentially. Formally, given a question $Q$, the LLM $\mathcal{M}_L$ generates the thinking insights accompanied with the prompt $p$:
\begin{equation}
    \mathcal{I} = \mathcal{M}_L(Q, p). \label{eq:llmI}
\end{equation}

\subsection{Cost-Aware Continual Judgment}
\label{ssec:judgment}

Problems of varying difficulty require differing levels of reasoning guidance. To avoid wasting computational resources on simple problems while ensuring sufficient support for more complex ones, we decompose the reasoning process into stages and introduce a cost-aware judgment mechanism. This mechanism evaluates whether the current reasoning content is sufficient for the SLM to generate a reliable response, allowing the LLM to terminate reasoning early and skip unnecessary stages.

\paragraph{Insight Cutoff.} To balance reasoning quality and computational cost, we divide the whole thinking process into several successive stages with increasing \textbf{effort levels}, e.g., three in this paper, namely low, medium, and high.
Each stage generates insights covering all four types (Goal, Planning, Retrieval, Action), but with progressively larger token budgets and greater depth of analysis.  
Let $\Delta\mathcal{I}^t$ denote the newly generated insights at stage $t$; the cumulative insight set is constructed as:
\begin{equation}
    \mathcal{I}^{0} = \emptyset, \quad \mathcal{I}^{t} = \mathcal{I}^{t-1} \oplus \Delta\mathcal{I}^{t} \quad (t = 1, 2, 3),
\end{equation}
where $\oplus$ denotes sequential concatenation of insights. $\mathcal{I}^{0} = \emptyset$ represents the baseline case where no reasoning insight is provided. The insight of the last stage $\mathcal{I}^{3}$ is equivalent to the full response $\mathcal{I}$ as defined in Equation~\eqref{eq:llmI}. This progressive design enables adaptive allocation of reasoning guidance: simple problems may terminate early at low effort (e.g., $\mathcal{I}^{1}$), while complex ones proceed to higher stages for richer support.

\paragraph{Confidence Measurement.}
To assess the sufficiency of the specific reasoning stage, we use the SLM's token-level probability distribution as the confidence signal. 
Previous studies have found that lower perplexity indicates the SLM finds insights linguistically familiar, while lower entropy reflects higher predictive stability~\cite{Uncertainty}.

Our empirical analysis, presented in Figure~\ref{fig:entropy_distribution}, also reveals a consistent pattern: the SLM produces lower output entropy under guidance that leads to correct answers—compared to incorrect ones—with a more pronounced separation observed at higher reasoning effort levels. This finding motivates the use of perplexity and entropy as diagnostic indicators of reasoning sufficiency.

Concretely, for each stage $t$, we concatenate $Q$ and $\mathcal{I}^t$ to form the input sequence $x^t$ for the SLM $\mathcal{M}_S$, i.e., $x^t = Q \oplus \mathcal{I}^t$. Let $x^t_i$ denote the $i$-th token in $x^t$; the corresponding predictive distribution at position $i$ is computed as:
\begin{equation}
    P^t_i = P_{\mathcal{M}_S}(\cdot \mid x^t_{<i}).
\end{equation}
From this distribution, we compute per-token perplexity and entropy:
\begin{align}
    \mathrm{PPL}^t_i &= \exp(-\log P^t_i(x^t_i)) \\
    \mathrm{H}^t_i &= -\sum_{v \in \mathcal{V}} P^t_i(v) \log P^t_i(v),
\end{align}
where $\mathcal{V}$ denotes the vocabulary.

\paragraph{Sufficiency Classification.} 

Leveraging the distributional discrepancy between sufficient and insufficient guidance in Figure~\ref{fig:entropy_distribution}, we construct a classifier to retroactively assess reasoning sufficiency based on distributional statistics.
Let $n$ denote the number of tokens in the SLM input $x^t$. 
We denote the per-token PPL and entropy sequences as:
\begin{equation}
    \mathbf{PPL}^t = \{\mathrm{PPL}^t_i\}_{i=1}^{n}, \quad \mathbf{H}^t = \{\mathrm{H}^t_i\}_{i=1}^{n}.
\end{equation}
Based on these sequences, we construct a distributional feature vector:
\begin{equation}
    \mathbf{f}^t = \left[ \phi(\mathbf{PPL}^t), \phi(\mathbf{H}^t), \Delta_{\mathrm{PPL}}, \Delta_{\mathrm{H}}, n \right],
\end{equation}
where $\phi(\cdot)$ computes distributional statistics (mean, standard deviation, median, max, min, 25th/75th percentiles), and $\Delta$ denotes trend indicators measuring the difference between the last 20 and first 20 tokens, capturing whether confidence increases or decreases as the model processes the insight. Details are provided in Appendix~\ref{app:features}.

A classifier $\mathcal{C}$, exemplified as a multilayer perceptron (MLP), is trained on these features with binary labels indicating whether the SLM produces a correct answer under the given guidance. At inference time, the classifier outputs:
\begin{equation}
    s^t = \mathcal{C}(\mathbf{f}^t) \in [0, 1],
\end{equation}
where $s^t$ denotes the sufficiency score. A binary decision is made as $y^t = \mathbf{1}[s^t > \tau^t]$, with threshold $\tau^t$ optimized per effort level on a held-out validation set. We also explore alternative input representations (e.g., hidden states) and whether fine-tuning the SLM benefits classification in Appendix~\ref{app:classifier}.

\subsection{Response Completion}
Guided by the sufficiency assessments from the judgment mechanism, the framework dynamically determines when to halt the LLM's reasoning and transition control to the SLM, which then continues the reasoning process and generates the final answer.
Specifically, at stage $t$, if the sufficiency decision $y^t = 1$, the SLM generates the final answer $A$ using the question and current insights:
\begin{equation}
    A = \mathcal{M}_S(Q \oplus \mathcal{I}^t).
\end{equation}

If all stages yield $y^t = 0$, we fall back to selecting the configuration with the highest sufficiency score:
\begin{equation}
    t^* = \argmax_{0\leq t \leq 3} s^t,
\end{equation}
where $s^0$ is computed by applying the same classifier $\mathcal{C}$ to distributional features extracted when the SLM processes $Q$ alone. The final answer is then generated as:
\begin{equation}
    A = \mathcal{M}_S(Q \oplus \mathcal{I}^{t^*}).
\end{equation}

\section{Experiments}
\label{sec:experiments}

To evaluate the effectiveness of \mname, the following research questions (RQs) are investigated:
\begin{itemize}[leftmargin=*]
    \item \textbf{RQ1:} How does \mname perform compared to a single LLM?
    \item \textbf{RQ2:} Does \mname generalize to collaborations between models from different families?
    \item \textbf{RQ3:} What are the scaling behaviors of LLM collaboration with respect to guidance length and large--small model size combinations?
    \item \textbf{RQ4:} Can \mname utilize API-accessible LLM?
    \item \textbf{RQ5:} Does \mname generalize beyond mathematical reasoning, and how does it compare with efficiency-focused baselines?
\end{itemize}

\subsection{Experiment Settings}
\label{ssec:experiments-settings}

\begin{figure}[t]
    \centering
    \includegraphics[width=\linewidth]{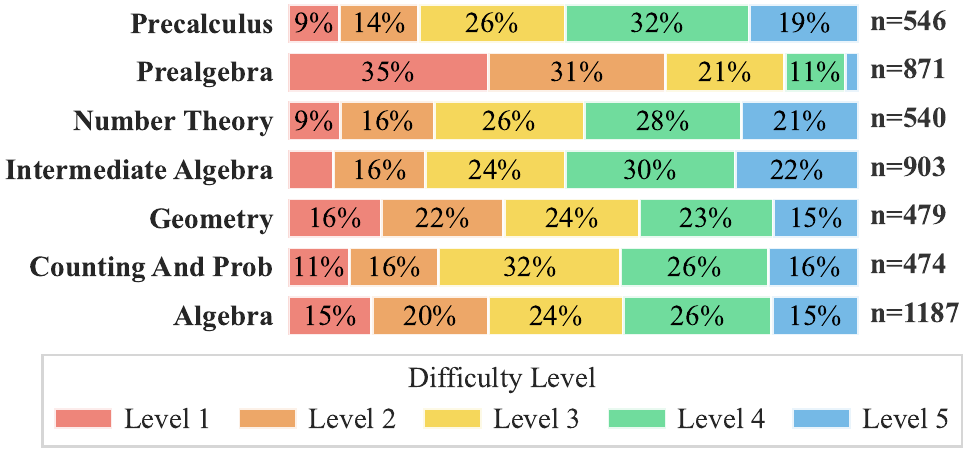}
    \caption{Sample distribution analysis of the MATH dataset~\cite{hen_math} across seven subjects and five difficulty levels. Difficulty levels range from 1 (easiest) to 5 (hardest).}
    \label{fig:dataset}
\end{figure}

\paragraph{Datasets.}

The experiments are conducted on two English mathematical reasoning benchmarks: MATH~\cite{hen_math} and GSM8K~\cite{gsm8k}. The MATH dataset contains 12.5K competition-level problems across seven subjects with five difficulty levels, split into 7.5K training and 5K test samples. The distribution of the test set is shown in Figure~\ref{fig:dataset}. GSM8K consists of 8.5K grade school math problems, split into 7.5K training and 1K test samples. Both datasets demand multi-step reasoning, making them suitable for evaluating \mname.

\paragraph{Evaluation Metrics.}

We evaluate model performance using three complementary metrics:
(1) \textbf{Accuracy}, defined as the percentage of correctly solved problems under the standard evaluation protocol of each dataset;
(2) \textbf{Inference Length}, defined as the total number of tokens generated during inference. Let $L_L$ and $L_S$ denote the numbers of tokens generated by the LLM and SLM, respectively; the overall inference length is $L_L + L_S$;
and (3) \textbf{Computational Cost}, approximated by the total TFLOPs incurred by both models. Let $|\theta_L|$ and $|\theta_S|$ denote the parameter counts of the LLM and SLM, respectively. The cost is computed as:
\begin{equation}
    \text{Cost}=\frac{2}{10^{12}}\left(|\theta_L|L_L+|\theta_S|(L_L+L_S)\right).
\end{equation}

% \begin{equation}
%     \text{Len} =  L_L + L_S
% \end{equation}

\begin{table*}[t]
\centering
\resizebox{\textwidth}{!}{
\begingroup
\setlength{\tabcolsep}{4pt}
\renewcommand{\arraystretch}{1.1}
\newcolumntype{Y}{>{\centering\arraybackslash}p{1.6cm}}
\begin{tabular}{@{}lllYYYYYYYYY@{}}
\toprule
\textbf{Series} & \textbf{Model} & \textbf{Metric} 
& \textbf{Algebra} 
& \textbf{Counting} 
& \textbf{Geometry} 
& \textbf{Intermediate} 
& \textbf{Num} 
& \textbf{Prealgebra} 
& \textbf{Precalc}
& \textbf{Average} \\ \midrule

\multirow{6}{*}{\textbf{Single}} 
 & \multirow{3}{*}{\textbf{7B}} 
 & Acc. & 93.09 & 75.95 & 65.34 & 60.13 & 76.85 & 89.55 & 62.45 & 77.14 \\
 &  & Len. & 2,194 & {\ul 2,786} & {\ul 2,854} & {\ul 3,218} & 2,891 & 2,084 & {\ul 3,098} & {\ul 2,732} \\
 &  & Cost & \textbf{30.71} & \textbf{39.00} & \textbf{39.96} & \textbf{45.06} & \textbf{40.47} & \textbf{29.17} & \textbf{43.37} & \textbf{38.25} \\ 
 \cmidrule(l){2-11}

 & \multirow{3}{*}{\textbf{32B}} 
 & Acc. & 94.78 & 82.07 & 68.89 & 64.45 & 85.00 & 91.04 & 67.22 & 80.90 \\
 &  & Len. & \textbf{2,064} & \textbf{2,676} & \textbf{2,816} & \textbf{3,162} & \textbf{2,678} & \textbf{2,013} & \textbf{3,003} & \textbf{2,630} \\
 &  & Cost & 132.13 & 171.26 & 180.26 & 202.36 & 171.40 & 128.84 & 192.20 & 168.35 \\ 
 \midrule

\multirow{9}{*}{\textbf{Baseline}} 
 & \multirow{3}{*}{\makecell[l]{\textbf{7B+32B} \\ \textbf{(low)}}} 
 & Acc. & 94.27 & 77.43 & 68.27 & 60.24 & 80.74 & 90.36 & 65.38 & 78.74 \\
 &  & Len. & {\ul 2,155} & 2,803 & 2,912 & 3,250 & {\ul 2,860} & {\ul 2,060} & 3,106 & 2,735 \\
 &  & Cost & {\ul 36.64} & {\ul 45.71} & {\ul 47.24} & {\ul 51.97} & {\ul 46.51} & {\ul 35.30} & {\ul 49.95} & {\ul 44.76} \\ 
 \cmidrule(l){2-11}

 & \multirow{3}{*}{\makecell[l]{\textbf{7B+32B} \\ \textbf{(medium)}}} 
 & Acc. & 93.93 & 79.96 & 69.94 & 63.90 & 83.33 & 91.16 & 67.40 & 80.36 \\
 &  & Len. & 2,205 & 2,959 & 3,034 & 3,380 & 2,963 & 2,116 & 3,312 & 2,853 \\
 &  & Cost & 62.90 & 73.49 & 74.48 & 79.39 & 73.53 & 61.53 & 78.43 & 71.96 \\ 
 \cmidrule(l){2-11}

 & \multirow{3}{*}{\makecell[l]{\textbf{7B+32B} \\ \textbf{(high)}}} 
 & Acc. & {\ul 95.53} & {\ul 82.49} & {\ul 72.86} & {\ul 67.55} & {\ul 88.52} & {\ul 92.54} & {\ul 71.61} & {\ul 83.18} \\
 &  & Len. & 2,178 & 3,057 & 3,190 & 3,580 & 2,980 & 2,119 & 3,406 & 2,930 \\
 &  & Cost & 93.87 & 106.77 & 108.39 & 114.07 & 105.47 & 92.10 & 111.64 & 104.62 \\ 
 \midrule

\multirow{6}{*}{\textbf{\mname}} 
 & \multirow{6}{*}{\textbf{7B+32B}} 
 & \multirow{2}{*}{Acc.} 
   & \textbf{96.04} & \textbf{82.70} & \textbf{73.07} & \textbf{67.77} 
   & \textbf{88.70} & \textbf{92.65} & \textbf{71.98} & \textbf{83.46} \\
 &  &  & \accpct{96.04}{94.78} & \accpct{82.70}{82.07} & \accpct{73.07}{68.89} & \accpct{67.77}{64.45} 
   & \accpct{88.70}{85.00} & \accpct{92.65}{91.04} & \accpct{71.98}{67.22} & \accpct{83.46}{80.90} \\

 &  & \multirow{2}{*}{Len.} 
   & 2,175 & 3,057 & 3,136 & 3,549 & 2,976 & 2,118 & 3,402 & 2,916 \\
 &  &  & \downpct{2175}{2064} & \downpct{3057}{2676} & \downpct{3136}{2816} & \downpct{3549}{3162} 
   & \downpct{2976}{2678} & \downpct{2118}{2013} & \downpct{3402}{3003} & \downpct{2916}{2630} \\

 &  & \multirow{2}{*}{Cost} 
   & 86.27 & 105.78 & 96.26 & 107.02 & 99.86 & 91.92 & 110.94 & 99.72 \\
 &  &  & \downpct{86.27}{132.13} & \downpct{105.78}{171.26} & \downpct{96.26}{180.26} & \downpct{107.02}{202.36} 
   & \downpct{99.86}{171.40} & \downpct{91.92}{128.84} & \downpct{110.94}{192.20} & \downpct{99.72}{168.35} \\
\bottomrule
\end{tabular}
\endgroup
}
\caption{Performance comparison of different model configurations on the MATH dataset~\cite{hen_math} using DeepSeek-7B and -32B. We report accuracy (\%), average inference length (tokens), and computational cost (TFLOPs). Bold values denote the best performance and underlined values denote the second-best performance for each subject. Percentage changes are computed relative to DeepSeek-32B.}
\label{tab:overall}
\end{table*}

\paragraph{Implementation details.}

For the \textbf{LLMs}, we consider two model families: DeepSeek-R1~\cite{deepseekr1} and Qwen3~\cite{i:18_Qwen3}. We use DeepSeek-R1-Distill-Qwen-7B and DeepSeek-R1-Distill-Qwen-32B (denoted as DeepSeek-7B and DeepSeek-32B), together with Qwen3-8B and Qwen3-32B. All LLMs operate in the thinking mode with deterministic generation settings: temperature = 0, top-p = 1.0, and no frequency penalty. To investigate the cost-performance trade-off, we empirically evaluate three fixed thinking lengths: 100, 500, and 1,000 tokens (corresponding to low, medium, and high effort levels). The maximum output length for complete answers is set to 8,192 tokens.

For the \textbf{continual classifier}, we train it on the training splits of both datasets; for MATH, we train a separate classifier for each subject. A sample is labeled as \emph{sufficient} ($y=1$) if the SLM produces the correct answer given the current insights, and as \emph{insufficient} ($y=0$) otherwise. We adopt a stratified 70/30 train--validation split with random seed 42. The classifier is a two-hidden-layer MLP (64 and 32 units) with ReLU activations and dropout ($p=0.3$), optimized with Adam (lr $=10^{-4}$) for up to 3 epochs, using early stopping based on validation accuracy. The threshold $\tau_e$ for each effort level is determined by grid search over the range $[0.05, 0.95]$ with step size $0.05$ on the training set.

\subsection{Overall Performance Comparison (RQ1)}
\label{ssec:rq1}

As shown in Table~\ref{tab:overall}, the experiment compares the results of single model inference, fixed-length thinking baselines, and our \mname approach across seven mathematical subjects.

\paragraph{Thinking Length Analysis.} Overall, incorporating the 32B model's initial thinking improves the 7B model's performance across all subjects. Notably, with sufficient guidance (high effort level), the collaboration not only surpasses the 7B baseline but also exceeds the 32B model's standalone performance on most subjects. This suggests a synergistic effect where the combination of the LLM's high-level reasoning and the SLM's execution capability achieves results beyond what either model can accomplish alone.

\paragraph{\mname Performance.} \mname effectively identifies beneficial thinking processes and adaptively allocates computational resources. It achieves the best accuracy across all subjects, outperforming both the 7B and 32B standalone models. Compared to the 32B mentor model, \mname achieves 2.56 percentage points higher accuracy while requiring only 59\% of the computational cost. These results demonstrate that our cost-aware judgment mechanism successfully balances performance and efficiency through dynamic resource allocation.

\begin{table}[t]
\centering
\resizebox{\linewidth}{!}{
\begin{tabular}{@{}lcccccc@{}}
\toprule
                                 & \multicolumn{3}{c}{\textbf{MATH}}                & \multicolumn{3}{c}{\textbf{GSM8K}}               \\ \cmidrule(l){2-7} 
\textbf{Model}                   & \textbf{Acc.}  & \textbf{Len.}  & \textbf{Cost}  & \textbf{Acc.}  & \textbf{Len.}  & \textbf{Cost}  \\ \midrule
\textit{Single Models}           &                &                &                &                &                &                \\
\quad \ding{192} Qwen3-8B                   & 60.86          & 3,197          & \underline{51.15}    & 89.61          & 1,991          & \underline{31.86}    \\
\quad \ding{193} Qwen3-32B                  & 69.50          & 3,022          & 193.41         & 94.01          & 1,625          & 104.00         \\
\quad \ding{194} DeepSeek-7B                & 76.92          & 2,661          & \textbf{37.25} & 87.11          & 1,124          & \textbf{15.74} \\
\quad \ding{195} DeepSeek-32B               & \underline{80.76}    & \underline{2,560}    & 163.84         & 94.47          & \textbf{1,049} & 67.14          \\ \midrule
\textit{Collaboration}           &                &                &                &                &                &                \\
\quad \ding{192}+\ding{193}       & 64.42          & 3,160          & 96.69          & 94.77          & 1,727          & 57.00          \\
\quad \ding{192}+\ding{195}      & 63.72          & 3,055          & 94.71          & \underline{95.22}    & 1,360          & 44.18          \\
\quad \ding{194}+\ding{193}      & 79.96          & \textbf{2,520} & 58.06          & 94.62          & 1,488          & 76.87          \\
\quad \ding{194}+\ding{195}   & \textbf{83.34} & 2,820          & 97.95          & \textbf{95.45} & \underline{1,059}    & 52.66          \\ \bottomrule
\end{tabular}
}
\caption{Performance of cross-family collaborations on the MATH~\cite{hen_math} and GSM8K~\cite{gsm8k} datasets. Bold and underlined values denote the best and second-best performance, respectively.}
\label{tab:cross_family}
\end{table}

\subsection{Cross-Family Generalization (RQ2)}
\label{ssec:rq2}

To evaluate whether \mname generalizes beyond a single model family, we test collaborations between DeepSeek and Qwen3. Table~\ref{tab:cross_family} presents results on both the MATH and GSM8K datasets.

\paragraph{Cross-family collaboration is effective.} Our experiments show that cross-family collaboration yields consistent performance gains over SLM-only baselines. For instance, DeepSeek-7B paired with Qwen3-32B achieves 79.96\% accuracy on MATH, surpassing both DeepSeek-7B alone (76.92\%) and Qwen3-32B alone (69.50\%), while reducing inference cost to less than one-third of using Qwen3-32B. This indicates that the structured insights generated by \mname are sufficiently general to be understood and utilized across different families.

\paragraph{Capability alignment matters.} The magnitude of improvement depends on the capability gap between models. When the SLM is relatively weak (e.g., Qwen3-8B), collaboration with DeepSeek-32B improves MATH accuracy by only 2.86\% and fails to reach the LLM's standalone performance. We attribute this limitation to a comprehension bottleneck: weaker SLMs may lack the capacity to fully interpret abstract reasoning guidance, causing them to underutilize or misapply the LLM's insights. This suggests that effective collaboration requires reasonable capability alignment between mentor and intern models.

\begin{figure}[t]
    \centering
    \includegraphics[width=\linewidth]{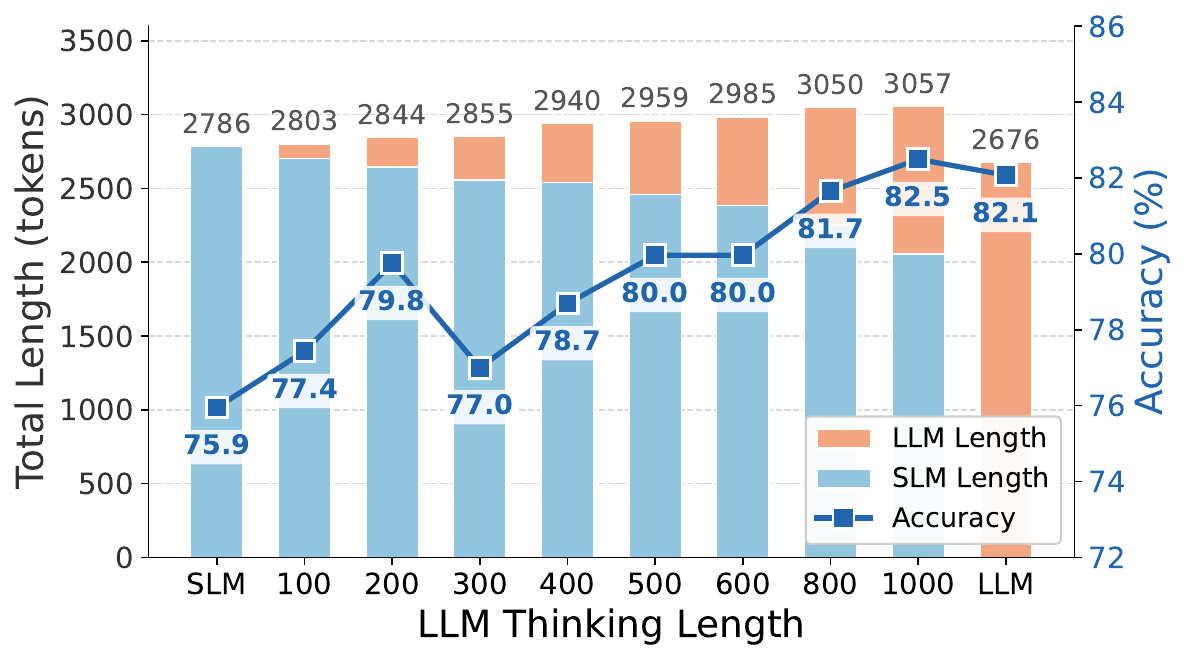}
    \caption{Accuracy and token length breakdown on the Counting \& Probability subset of MATH~\cite{hen_math}. The stacked bars show the composition of total length (SLM in blue, LLM in orange), while the line indicates accuracy.}
    \label{fig:length}
\end{figure}

\subsection{Scaling Law of LLM Collaboration (RQ3)}
\label{ssec:rq3}

\paragraph{Guidance Length.}

To understand how guidance length impacts performance, we analyze the trade-off between total token length and accuracy when using DeepSeek-32B and -7B. As shown in Figure~\ref{fig:length}, accuracy generally improves as guidance length increases, eventually surpassing the pure LLM baseline. Notably, even with minimal guidance (200 tokens), the collaboration substantially outperforms the SLM-only baseline. We attribute this to the structured nature of insights: even brief guidance provides goal clarification and high-level planning that helps the SLM avoid dead-ends.

However, the performance curve exhibits fluctuations across different guidance lengths, suggesting that the optimal amount of guidance is problem-dependent. Simple problems may be solved with brief hints, while complex ones require detailed action steps. This observation motivates our cost-aware judgment mechanism, which dynamically determines the appropriate level of guidance rather than relying on a fixed thinking budget.

\begin{table}[t]
\centering
\resizebox{\linewidth}{!}{
\begin{tabular}{@{}cccccccc@{}}
\toprule
\multirow{2}{*}{\textbf{SLM}} & \multirow{2}{*}{\textbf{LLM}} & \multicolumn{2}{c}{\textbf{Low}} & \multicolumn{2}{c}{\textbf{Medium}} & \multicolumn{2}{c}{\textbf{High}} \\
\cmidrule(lr){3-4} \cmidrule(lr){5-6} \cmidrule(lr){7-8}
 &  & Acc. & Len. & Acc. & Len. & Acc. & Len. \\ \midrule
1.5B & --   & 59.92 & 2,898 & \multicolumn{2}{c}{--} & \multicolumn{2}{c}{--} \\
1.5B & 7B   & 59.28 & 3,051 & 58.23 & 3,446 & 60.55 & 3,925 \\
1.5B & 14B  & 61.13 & 2,961 & 63.50 & 2,866 & 65.82 & 2,801 \\
1.5B & 32B  & 60.97 & 2,956 & 62.66 & 2,877 & 67.09 & 2,854 \\ \midrule
7B   & --   & 75.95 & 2,786 & \multicolumn{2}{c}{--} & \multicolumn{2}{c}{--} \\
7B   & 14B  & 74.68 & 2,685 & 75.74 & 2,669 & 75.11 & 2,690 \\
7B   & 32B  & 77.43 & 2,803 & 79.96 & 2,959 & 82.49 & 3,057 \\ \midrule
14B  & --   & 79.96 & 2,613 & \multicolumn{2}{c}{--} & \multicolumn{2}{c}{--} \\
14B  & 32B  & 80.17 & 2,581 & 79.48 & 2,628 & 80.38 & 2,665 \\ \midrule
32B  & --   & 82.07 & 2,675 & \multicolumn{2}{c}{--} & \multicolumn{2}{c}{--} \\
\bottomrule
\end{tabular}
}
\caption{Performance of LLM collaboration with different model size combinations on Counting and Probability with DeepSeek family. Rows with ``--'' as LLM denote single-model baselines. Low, Medium, and High correspond to the effort levels of LLM guidance.}
\label{tab:combination}
\end{table}

\paragraph{Model Size.}
To examine the role of model size, we evaluate different combinations of LLM and SLM sizes. Table~\ref{tab:combination} reports results ranging from 1.5B to 32B across three effort levels. Rows with ``--'' as LLM denote single-model inference.

A key finding is that effective collaboration requires a moderate capability gap between the LLM and SLM. When the gap is too large, the weaker SLM shows only modest improvements: extremely small models may lack sufficient capacity to parse and execute complex reasoning patterns from much larger models, leading to partial utilization of the guidance. For instance, the 1.5B model with 32B guidance only improves from 59.92\% to 67.09\%, failing to match the 32B standalone performance of 82.07\%. Conversely, when the gap is too small, collaboration yields marginal gains, possibly because models of similar capacity share similar reasoning styles and thus provide redundant rather than complementary insights. The optimal collaboration emerges when the SLM is capable enough to follow the LLM's guidance yet still benefits from higher-level reasoning it cannot produce independently.

\subsection{API-accessible LLM Collaboration (RQ4)}
\label{ssec:rq4}

\begin{table}[t]
\centering
\resizebox{\linewidth}{!}{
\begin{tabular}{@{}lcccccc@{}}
\toprule
\multirow{2}{*}{\textbf{Model}} & \multicolumn{2}{c}{\textbf{Algebra}} & \multicolumn{2}{c}{\textbf{Counting}} & \multicolumn{2}{c}{\textbf{Geometry}} \\ 
\cmidrule(lr){2-3} \cmidrule(lr){4-5} \cmidrule(lr){6-7}
 & Acc. & Cost & Acc. & Cost & Acc. & Cost \\ \midrule
\textit{Single Models} & & & & & & \\
\quad \ding{192} DeepSeek-7B & 93.09 & 0.18 & 74.26 & \textbf{0.22} & 65.34 & \textbf{0.23} \\
\quad \ding{193} GPT-4o-mini & 91.32 & 0.31 & 74.89 & 0.37 & 62.42 & 0.40 \\
\quad \ding{194} GPT-oss-120B & 82.14 & {\ul 0.19} & {\ul 86.50} & 0.29 & {\ul 82.05} & 0.50 \\ \midrule
\textit{Collaboration} & & & & & & \\
\quad \ding{192}+\ding{193} & {\ul 94.10} & 0.17 & 86.29 & 0.77 & 77.66 & 0.31 \\
\quad \ding{192}+\ding{194} & \textbf{95.79} & \textbf{0.10} & \textbf{86.71} & {\ul 0.23} & \textbf{84.55} & {\ul 0.30} \\
\bottomrule
\end{tabular}
}
\caption{Performance and cost (\$/1K samples) of collaboration with API-based LLMs on selected MATH subjects. DeepSeek-7B serves as the local SLM, while GPT-4o-mini and gpt-oss-120b are accessed via API.}
\label{tab:api}
\end{table}

A practical advantage of \mname is its compatibility with closed-source LLMs accessible only via API, as it does not require access to model weights. To validate this, we evaluate collaborations where DeepSeek-7B serves as the local SLM, while GPT-4o-mini~\cite{openai2024gpt4ocard} and gpt-oss-120b~\cite{oss} are accessed through API calls.

As shown in Table~\ref{tab:api}, API-based collaboration consistently improves accuracy across all subjects, with the combined system outperforming both the local SLM and the remote LLM alone. This confirms that \mname remains effective even when the mentor is accessed via API, likely because the structured insight format transfers well regardless of communication protocol.

Notably, the collaboration also reduces inference cost. The LLM's high-level thinking insights enable the SLM to solve problems with shorter reasoning chains, eliminating the need for lengthy self-exploration. This cost reduction is particularly valuable for API-based deployment, where pricing is typically token-based and verbose outputs directly increase expenses.

\subsection{Cross-Domain Generalization and Efficiency Baselines (RQ5)}
\label{ssec:rq5}

\begin{table}[t]
\centering
\resizebox{\linewidth}{!}{
\begin{tabular}{@{}lcc@{}}
\toprule
\textbf{Method} & \textbf{HumanEval Acc.} & \textbf{Correct / Total} \\ \midrule
\textit{Single Models} & & \\
\quad SLM (7B) & 65.24 & 107 / 164 \\
\quad LLM (32B) & 89.02 & 146 / 164 \\ \midrule
\textit{Collaboration} & & \\
\quad 7B+32B (low) & 73.17 & 120 / 164 \\
\quad 7B+32B (medium) & 79.27 & 130 / 164 \\
\quad 7B+32B (high) & {\ul 83.54} & 137 / 164 \\
\quad \mname & \textbf{85.37} & 140 / 164 \\ \bottomrule
\end{tabular}
}
\caption{Cross-domain evaluation on HumanEval~\cite{humaneval} (code generation, 164 problems). The sufficiency classifier is trained on MATH and applied to HumanEval without retraining.}
\label{tab:humaneval}
\end{table}

\begin{table}[t]
\centering
\resizebox{\linewidth}{!}{
\begin{tabular}{@{}lcc@{}}
\toprule
\textbf{Method} & \textbf{MATH Acc. (\%)} & \textbf{Cost (TFLOPs)} \\ \midrule
SLM (7B) & 77.14 & \textbf{38.25} \\
LLM (32B) & 80.90 & 168.35 \\
Budget Forcing~\cite{muennighoff-etal-2025-s1} & 82.18 & 108.74 \\
LLM Cascade~\cite{FrugalGPT} & {\ul 82.60} & {\ul 95.33} \\
\mname & \textbf{83.46} & 99.72 \\ \bottomrule
\end{tabular}
}
\caption{Comparison with efficiency-focused baselines on MATH. Budget Forcing truncates the 32B model's reasoning at a fixed token budget. LLM Cascade makes a one-time binary routing decision between SLM and 7B+32B (high).}
\label{tab:efficiency}
\end{table}

\paragraph{Cross-domain generalization.} To evaluate whether \mname generalizes beyond mathematical reasoning, we test on HumanEval~\cite{humaneval}, a code generation benchmark of 164 programming problems. Crucially, we apply the sufficiency classifier trained on MATH directly to HumanEval \emph{without any retraining}. As shown in Table~\ref{tab:humaneval}, the collaboration pattern holds: accuracy improves monotonically with guidance level, and \mname achieves 85.37\%, surpassing the best fixed-budget baseline (83.54\%) by 1.83\%. This confirms that the PPL/entropy features capture domain-agnostic confidence signals, enabling effective cross-domain transfer.

\paragraph{Comparison with efficiency baselines.} We compare \mname with two representative efficiency-focused methods on MATH. Following FrugalGPT~\cite{FrugalGPT}, we implement an LLM Cascade that routes each sample to either SLM alone or 7B+32B (high) via a binary classifier. Following s1~\cite{s1}, Budget Forcing truncates the 32B model's reasoning at a fixed token budget. As shown in Table~\ref{tab:efficiency}, \mname achieves the highest accuracy (83.46\%) while maintaining competitive cost. Compared to LLM Cascade, \mname gains +0.86\% accuracy through per-stage continual judgment rather than a one-time routing decision. Compared to Budget Forcing, \mname is both more accurate (+1.28\%) and more efficient (8.3\% lower cost), because the LLM generates compact structured insights rather than full reasoning chains.
Additional analysis of classifier accuracy, stage distribution, latency measurements, and failure cases is provided in Appendix~\ref{app:analysis}.

\section{Related Works}
\label{related-works}

LLM collaboration methods leverage multiple models working together to solve complex problems through division of labor and iterative discussion.
These approaches can be broadly categorized by their coordination mechanisms.
Debate-based methods~\cite{ChatEval} let models discuss and refine responses through multi-turn exchanges, where each model critiques and improves upon others' outputs until consensus is reached.
Role-based methods~\cite{MetaGPT} assign specialized functions to different models within structured workflows, such as planner, executor, and reviewer, enabling complex task decomposition.
Verification-based methods~\cite{Verify} employ one model to generate candidates while another validates or ranks them, improving output reliability through cross-model checking.
Recent work has also explored large--small model collaboration for inference acceleration, where smaller models assist larger ones through speculative decoding~\cite{Speculative} or draft-then-verify pipelines, reducing latency by parallelizing token generation.
However, most existing approaches either route each query to a single model without leveraging complementary capabilities, or combine multiple models at the cost of substantial overhead, leaving the challenge of achieving both improved quality and cost efficiency largely unaddressed.
A more comprehensive discussion of the broader landscape of LLM ensemble methods is provided in Appendix~\ref{app:ensemble}.

\section{Conclusion}
\label{sec:conclusion}

In this paper, we proposed \mname, a large--small LLM collaboration framework, which reduces computational cost while preserving the benefits of the thinking paradigm of LLMs. It separates reasoning from answer generation by letting LLMs provide structured insights and using a cost-aware mechanism to decide when these insights are sufficient for SLMs to produce answers. Experiments on mathematical reasoning and code generation benchmarks show that \mname outperforms the mentor LLM while reducing computation by roughly 40\%, and the sufficiency classifier transfers across domains without retraining. These results suggest that lightweight high-quality reasoning guidance can often serve as an effective substitute for full reasoning chains.

\section*{Limitations}

Our work has several limitations that suggest directions for future research.

First, while we have extended our evaluation to code generation (HumanEval) in addition to mathematical reasoning (MATH and GSM8K), the generalizability to other domains such as commonsense reasoning or open-ended question answering remains to be further explored.

Second, although our experiments show that a unified classifier and cross-domain transfer are viable (Appendix~\ref{app:unified}), the classifier still requires labeled training data from at least one domain. Reducing this supervision requirement is an interesting direction.

Third, \mname employs a fixed two-model collaboration where one LLM serves as the mentor and one SLM as the intern. This simple division may not fully exploit more sophisticated collaboration patterns, such as involving multiple models with diverse capabilities or dynamically adjusting roles based on task characteristics. Exploring finer-grained multi-model collaboration could further enhance performance.

\section*{Acknowledgments}

This research was partially supported by National Natural Science Foundation of China (No.62502404), Hong Kong Research Grants Council (Research Impact Fund No.R1015-23, Collaborative Research Fund No.C1043-24GF, General Research Fund No. 11218325), Institute of Digital Medicine of City University of Hong Kong (No.9229503), and Tencent (Tencent Rhino-Bird Focused Research Program, Tencent University Cooperation Project).

\bibliography{custom}

\appendix

\begin{table*}[t]
\centering
\resizebox{\textwidth}{!}{
\begin{tabular}{@{}lllrlrlrlrlrlrlrlrl@{}}
\toprule
\textbf{Series} & \textbf{Model} & \textbf{Metric} 
& \multicolumn{2}{c}{\textbf{Algebra}} 
& \multicolumn{2}{c}{\textbf{Counting}} 
& \multicolumn{2}{c}{\textbf{Geometry}} 
& \multicolumn{2}{c}{\textbf{Intermediate}} 
& \multicolumn{2}{c}{\textbf{Num}} 
& \multicolumn{2}{c}{\textbf{Prealgebra}} 
& \multicolumn{2}{c}{\textbf{Precalc}}
& \multicolumn{2}{c}{\textbf{Average}} \\ \midrule

\multicolumn{19}{c}{\textbf{Non-thinking Mode}} \\ \midrule

\multirow{6}{*}{\textbf{Single}} 
 & \multirow{3}{*}{\textbf{7B}} 
 & Acc. 
 & \underline{93.60} &  & 76.79 &  & 67.43 &  & 67.33 &  & \underline{85.74} &  & 89.55 &  & 67.95 &  & 80.40 &  \\
 &  & Len. 
 & 487.0 &  & 542.2 &  & \textbf{693.5} &  & \underline{1101.0} &  & 640.5 &  & 370.8 &  & \textbf{900.4} &  & 664.4 &  \\
 &  & Cost 
 & \textbf{6.82} &  & \textbf{7.59} &  & \textbf{9.71} &  & \textbf{15.41} &  & \textbf{8.97} &  & \textbf{5.19} &  & \textbf{12.61} &  & \textbf{9.47} &  \\ \cmidrule(l){2-19}

 & \multirow{3}{*}{\textbf{32B}} 
 & Acc. 
 & 92.84 & \accpct{92.84}{93.60} 
 & \underline{78.27} & \accpct{78.27}{76.79} 
 & 69.31 & \accpct{69.31}{67.43} 
 & 66.11 & \accpct{66.11}{67.33} 
 & 82.41 & \accpct{82.41}{85.74} 
 & \underline{90.59} & \accpct{90.59}{89.55} 
 & 66.48 & \accpct{66.48}{67.95} 
 & 79.98 & \accpct{79.98}{80.40} \\
 &  & Len. 
 & \textbf{385.5} & \downpct{385.5}{487.0} 
 & 452.1 & \downpct{452.1}{542.2} 
 & 839.8 & \downpct{839.8}{693.5} 
 & 1296.9 & \downpct{1296.9}{1101.0} 
 & \underline{555.4} & \downpct{555.4}{640.5} 
 & 319.2 & \downpct{319.2}{370.8} 
 & 1105.5 & \downpct{1105.5}{900.4} 
 & 685.3 & \downpct{685.3}{664.4} \\
 &  & Cost 
 & 24.67 & \downpct{24.67}{6.82} 
 & 28.93 & \downpct{28.93}{7.59} 
 & 53.75 & \downpct{53.75}{9.71} 
 & 83.00 & \downpct{83.00}{15.41} 
 & 35.55 & \downpct{35.55}{8.97} 
 & 20.43 & \downpct{20.43}{5.19} 
 & 70.75 & \downpct{70.75}{12.61} 
 & 45.30 & \downpct{45.30}{9.47} \\ \midrule

\multirow{9}{*}{\textbf{Baseline}} 
 & \multirow{3}{*}{\makecell[l]{\textbf{7B+32B} \\ \textbf{(low)}}} 
 & Acc. 
 & 91.83 & \accpct{91.83}{93.60} 
 & 75.53 & \accpct{75.53}{76.79} 
 & 67.43 & \accpct{67.43}{67.43} 
 & 65.34 & \accpct{65.34}{67.33} 
 & 80.00 & \accpct{80.00}{85.74} 
 & 90.36 & \accpct{90.36}{89.55} 
 & 67.77 & \accpct{67.77}{67.95} 
 & 79.00 & \accpct{79.00}{80.40} \\
 &  & Len. 
 & 402.7 & \downpct{402.7}{487.0} 
 & 467.0 & \downpct{467.0}{542.2} 
 & \underline{699.0} & \downpct{699.0}{693.5} 
 & \textbf{1094.8} & \downpct{1094.8}{1101.0} 
 & 578.1 & \downpct{578.1}{640.5} 
 & 311.8 & \downpct{311.8}{370.8} 
 & \underline{966.1} & \downpct{966.1}{900.4} 
 & \underline{626.8} & \downpct{626.8}{664.4} \\
 &  & Cost 
 & 12.10 & \downpct{12.10}{6.82} 
 & \underline{13.00} & \downpct{13.00}{7.59} 
 & \underline{16.25} & \downpct{16.25}{9.71} 
 & \underline{21.79} & \downpct{21.79}{15.41} 
 & 14.56 & \downpct{14.56}{8.97} 
 & \underline{10.82} & \downpct{10.82}{5.19} 
 & \underline{19.98} & \downpct{19.98}{12.61} 
 & \underline{15.50} & \downpct{15.50}{9.47} \\ \cmidrule(l){2-19}

 & \multirow{3}{*}{\makecell[l]{\textbf{7B+32B} \\ \textbf{(medium)}}} 
 & Acc. 
 & 93.26 & \accpct{93.26}{93.60} 
 & 77.85 & \accpct{77.85}{76.79} 
 & 66.60 & \accpct{66.60}{67.43} 
 & 66.00 & \accpct{66.00}{67.33} 
 & 81.11 & \accpct{81.11}{85.74} 
 & 90.24 & \accpct{90.24}{89.55} 
 & \underline{70.70} & \accpct{70.70}{67.95} 
 & 80.02 & \accpct{80.02}{80.40} \\
 &  & Len. 
 & 392.3 & \downpct{392.3}{487.0} 
 & \textbf{416.2} & \downpct{416.2}{542.2} 
 & 774.9 & \downpct{774.9}{693.5} 
 & 1113.1 & \downpct{1113.1}{1101.0} 
 & \textbf{551.7} & \downpct{551.7}{640.5} 
 & 304.7 & \downpct{304.7}{370.8} 
 & 970.3 & \downpct{970.3}{900.4} 
 & \textbf{626.5} & \downpct{626.5}{664.4} \\
 &  & Cost 
 & 26.43 & \downpct{26.43}{6.82} 
 & 27.88 & \downpct{27.88}{7.59} 
 & 36.04 & \downpct{36.04}{9.71} 
 & 44.43 & \downpct{44.43}{15.41} 
 & 32.43 & \downpct{32.43}{8.97} 
 & 21.92 & \downpct{21.92}{5.19} 
 & 41.07 & \downpct{41.07}{12.61} 
 & 32.89 & \downpct{32.89}{9.47} \\ \cmidrule(l){2-19}

 & \multirow{3}{*}{\makecell[l]{\textbf{7B+32B} \\ \textbf{(high)}}} 
 & Acc. 
 & 93.34 & \accpct{93.34}{93.60} 
 & 77.43 & \accpct{77.43}{76.79} 
 & \underline{70.35} & \accpct{70.35}{67.43} 
 & \underline{67.77} & \accpct{67.77}{67.33} 
 & 83.89 & \accpct{83.89}{85.74} 
 & 90.59 & \accpct{90.59}{89.55} 
 & 69.78 & \accpct{69.78}{67.95} 
 & \underline{80.94} & \accpct{80.94}{80.40} \\
 &  & Len. 
 & \underline{390.1} & \downpct{390.1}{487.0} 
 & \underline{450.3} & \downpct{450.3}{542.2} 
 & 816.0 & \downpct{816.0}{693.5} 
 & 1167.3 & \downpct{1167.3}{1101.0} 
 & 558.7 & \downpct{558.7}{640.5} 
 & \underline{301.9} & \downpct{301.9}{370.8} 
 & 1039.4 & \downpct{1039.4}{900.4} 
 & 650.8 & \downpct{650.8}{664.4} \\
 &  & Cost 
 & 28.82 & \downpct{28.82}{6.82} 
 & 32.61 & \downpct{32.61}{7.59} 
 & 44.02 & \downpct{44.02}{9.71} 
 & 60.80 & \downpct{60.80}{15.41} 
 & 38.71 & \downpct{38.71}{8.97} 
 & 22.86 & \downpct{22.86}{5.19} 
 & 54.44 & \downpct{54.44}{12.61} 
 & 40.32 & \downpct{40.32}{9.47} \\ \midrule

\multirow{3}{*}{\textbf{\mname}} 
 & \multirow{3}{*}{\textbf{7B+32B}} 
 & Acc. 
 & \textbf{94.19} & \accpct{94.19}{93.60} 
 & \textbf{81.01} & \accpct{81.01}{76.79} 
 & \textbf{70.56} & \accpct{70.56}{67.43} 
 & \textbf{69.66} & \accpct{69.66}{67.33} 
 & \textbf{85.93} & \accpct{85.93}{85.74} 
 & \textbf{90.93} & \accpct{90.93}{89.55} 
 & \textbf{73.81} & \accpct{73.81}{67.95} 
 & \textbf{82.56} & \accpct{82.56}{80.40} \\
 &  & Len. 
 & 474.8 & \downpct{474.8}{487.0} 
 & 479.3 & \downpct{479.3}{542.2} 
 & 809.4 & \downpct{809.4}{693.5} 
 & 1118.6 & \downpct{1118.6}{1101.0} 
 & 639.3 & \downpct{639.3}{640.5} 
 & \textbf{301.9} & \downpct{301.9}{370.8} 
 & 1055.3 & \downpct{1055.3}{900.4} 
 & 674.6 & \downpct{674.6}{664.4} \\
 &  & Cost 
 & \underline{11.08} & \downpct{11.08}{6.82} 
 & 26.08 & \downpct{26.08}{7.59} 
 & 38.90 & \downpct{38.90}{9.71} 
 & 41.84 & \downpct{41.84}{15.41} 
 & \underline{9.40} & \downpct{9.40}{8.97} 
 & 21.54 & \downpct{21.54}{5.19} 
 & 51.91 & \downpct{51.91}{12.61} 
 & 28.68 & \downpct{28.68}{9.47} \\ 
\bottomrule
\end{tabular}
}
\caption{Performance comparison under non-thinking mode of the LLMs on the MATH dataset~\cite{hen_math}. We evaluate single models (7B, 32B), fixed-budget collaboration baselines (low/medium/high mentor token budgets), and \mname. Metrics include accuracy (\%), average inference length (tokens), and computational cost (TFLOPs). Bold indicates the best and underlined indicates the second-best for each subject.}
\label{tab:non-thinking}
\end{table*}

\section{Statistical Feature Extraction Details}
\label{app:features}

For the cost-aware thinking judgment mechanism described in Section~\ref{ssec:judgment}, we extract a comprehensive set of statistical features from the perplexity and entropy values computed over the insight tokens. These features form a feature vector $\mathbf{f}_e$ at each effort level $e$.

\paragraph{Perplexity Features (7 dimensions).}
Given $\mathbf{PPL}^t=\{\mathrm{PPL}^t_i\}_{i=1}^{n}$, we compute:
\begin{itemize}[leftmargin=*, itemsep=0pt]
  \item Mean: $\mu^{t}_{\mathrm{PPL}}=\frac{1}{n}\sum_{i=1}^{n}\mathrm{PPL}^{t}_{i}$
  \item Std:  $\sigma^{t}_{\mathrm{PPL}}=\sqrt{\frac{1}{n}\sum_{i=1}^{n}\left(\mathrm{PPL}^{t}_{i}-\mu^{t}_{\mathrm{PPL}}\right)^2}$
  \item Median: $\operatorname{median}(\{\mathrm{PPL}^{t}_{i}\}_{i=1}^{n})$
  \item Maximum: $\max(\{\mathrm{PPL}^{t}_{i}\}_{i=1}^{n})$
  \item Minimum: $\min(\{\mathrm{PPL}^{t}_{i}\}_{i=1}^{n})$
  \item 25th percentile: $Q_{25}(\{\mathrm{PPL}^{t}_{i}\}_{i=1}^{n})$
  \item 75th percentile: $Q_{75}(\{\mathrm{PPL}^{t}_{i}\}_{i=1}^{n})$
\end{itemize}

\paragraph{Entropy Features (7 dimensions).}
Given $\mathbf{H}^t=\{\mathrm{H}^t_i\}_{i=1}^{n}$, we compute:
\begin{itemize}[leftmargin=*, itemsep=0pt]
  \item Mean: $\mu^{t}_{\mathrm{H}}=\frac{1}{n}\sum_{i=1}^{n}\mathrm{H}^{t}_{i}$
  \item Std:  $\sigma^{t}_{\mathrm{H}}=\sqrt{\frac{1}{n}\sum_{i=1}^{n}\left(\mathrm{H}^{t}_{i}-\mu^{t}_{\mathrm{H}}\right)^2}$
  \item Median: $\operatorname{median}(\{\mathrm{H}^{t}_{i}\}_{i=1}^{n})$
  \item Maximum: $\max(\{\mathrm{H}^{t}_{i}\}_{i=1}^{n})$
  \item Minimum: $\min(\{\mathrm{H}^{t}_{i}\}_{i=1}^{n})$
  \item 25th percentile: $Q_{25}(\{\mathrm{H}^{t}_{i}\}_{i=1}^{n})$
  \item 75th percentile: $Q_{75}(\{\mathrm{H}^{t}_{i}\}_{i=1}^{n})$
\end{itemize}

\paragraph{Trend Features (2 dimensions).}
Let $k=\min(20,n)$. We compute:
\begin{itemize}[leftmargin=*, itemsep=0pt]
  \item Perplexity trend:
  $\Delta^{t}_{\mathrm{PPL}}=\frac{1}{k}\sum_{i=n-k+1}^{n}\mathrm{PPL}^{t}_{i}-\frac{1}{k}\sum_{i=1}^{k}\mathrm{PPL}^{t}_{i}$
  \item Entropy trend:
  $\Delta^{t}_{\mathrm{H}}=\frac{1}{k}\sum_{i=n-k+1}^{n}\mathrm{H}^{t}_{i}-\frac{1}{k}\sum_{i=1}^{k}\mathrm{H}^{t}_{i}$
\end{itemize}
These features capture whether the SLM becomes more or less confident as it processes $x^t$.

\paragraph{Metadata (1 dimension).}
\begin{itemize}[leftmargin=*, itemsep=0pt]
  \item \textbf{Length:} $n=|x^t|$, the total number of tokens in the SLM input $x^t=Q\oplus \mathcal{I}^t$.
\end{itemize}

These 18 features collectively provide a comprehensive representation of how the SLM processes the LLM-generated insights, capturing not only the overall difficulty but also the variability and temporal dynamics of the processing.

\begin{figure*}[t]
\centering
\begin{tcolorbox}[
    enhanced,
    width=0.95\linewidth,
    colback=white,
    colframe=black!70,
    boxrule=0.5pt,
    arc=2pt,
    left=8pt, right=8pt, top=6pt, bottom=6pt,
    fontupper=\small
]

\textcolor{sectionblue}{\textbf{\textsf{SYSTEM}}}\\[3pt]
You are a reasoning assistant. Analyze the given problem and follow the structured \texttt{[Thinking Insights]} below.

\vspace{6pt}
\textcolor{sectionblue}{\textbf{\textsf{INSIGHTS}}}
\begin{enumerate}[leftmargin=1.5em, itemsep=1pt, topsep=2pt, label=\arabic*.]
    \item \textbf{Goal}: $\langle$\textit{objective, constraints, and required output form}$\rangle$
    \item \textbf{Planning}: $\langle$\textit{high-level strategy; subproblem decomposition; edge cases}$\rangle$
    \item \textbf{Retrieval}: $\langle$\textit{relevant facts, formulas, or definitions; N/A if none}$\rangle$
    \item \textbf{Action}: $\langle$\textit{concrete steps and intermediate calculations}$\rangle$
\end{enumerate}

\vspace{4pt}
\textcolor{sectionblue}{\textbf{\textsf{CONSTRAINTS}}}
\begin{itemize}[leftmargin=1.5em, itemsep=1pt, topsep=2pt, label=\textbullet]
    \item Keep each component concise.
    \item Maintain notational consistency with the original problem.
\end{itemize}

\end{tcolorbox}
\caption{Prompt template for the LLM to generate structured Thinking Insights.}
\label{fig:prompt_schema}
\end{figure*}

\section{Broader Landscape of LLM Ensemble}
\label{app:ensemble}

LLM ensemble plays an essential part in enhancing the generation performance and robustness of AI systems~\cite{survey}. Existing LLM ensemble methods can be classified into three categories: 1) model selection before inference selects the best candidate LLM for a specific task or question, prioritizing smaller models whenever possible to minimize computational resources and inference time \cite{hybrid, i_5_before_inference_method_2_Zooter, forc, i:7_before_inference_method_4_Eagle, tf-llm-merging, mm-knowledge-editing, mta-personalized}; 2) Output aggregation methods obtain the best answer by integrating LLM outputs with different granularities. These methods can fully leverage the complementary capabilities of multiple LLMs when answering individual questions, thus offering higher potential performance \cite{gac, UniTe, i_10_during_inference_method_3_Citer, SweetSpan, i:12_during_inference_method_5_LE-MCTS, llmemb, plate-multiscene}; 3) LLM collaboration (i.e. LLM agent) employs multiple models working cooperatively through planning, reasoning and summary to reach a final answer. This approach can achieve the advantages of both previous categories through cascaded reasoning that starts with smaller models and escalates to larger ones based on output quality assessment \cite{Li2024MoreAI, LLM-Blender, i:15_collaboration_method_3_EcoAssistant, llm4rerank, llm-user-simulator, llm-multi-domain-ctr}.

\begin{table*}[t]
\centering
\resizebox{\linewidth}{!}{
\begin{tabular}{@{}lcccccccc@{}}
\toprule
                        & \textbf{Algebra} & \textbf{Counting} & \textbf{Geometry} & \textbf{Intermediate} & \textbf{Number} & \textbf{Prealgebra} & \textbf{Precalculus} & \textbf{Average} \\ \midrule
\textbf{SLM}            & 93.09            & 75.95             & 65.34             & 60.13                 & 76.85           & 89.55               & 62.45                & 71.74            \\
\textbf{LLM}            & 94.78            & 82.07             & 68.89             & 64.45                 & 85.00           & 91.04               & 67.22                & 80.90            \\
\textbf{Hidden}         & \underline{95.70} & \textbf{83.12}    & \underline{73.05} & \textbf{67.88}        & \underline{88.52} & \textbf{93.00}      & \underline{71.79}    & \underline{83.46} \\
\textbf{Hidden w/ LoRA} & 94.09            & 73.02             & 57.36             & 63.09                 & 75.60           & 91.45               & 69.57                & 74.88            \\ 
\textbf{Statistics}     & \textbf{96.04}   & \underline{82.70} & \textbf{73.07}    & \underline{67.77}     & \textbf{88.70}  & \underline{92.65}   & \textbf{71.98}       & \textbf{83.47}   \\
\bottomrule
\end{tabular}
}
\caption{Accuracy comparison across MATH subjects using different sufficiency prediction methods. All methods use MLP classifiers with different inputs: \textit{Statistics} uses perplexity and entropy from output probabilities; \textit{Hidden} uses last-layer hidden state at the final token; \textit{Hidden w/ LoRA} uses hidden states from a LoRA-finetuned SLM. Bold values indicate the best performance and underlined values indicate the second-best performance.}
\label{tab:classifier}
\end{table*}

\begin{table}[ht]
\centering
\begin{tabular}{@{}lcccc@{}}
\toprule
\textbf{Stage} & \textbf{Prec.} & \textbf{Recall} & \textbf{F1} \\ \midrule
SLM (7B) & 0.818 & 0.707 & 0.759 \\
7B+32B (low) & 0.828 & 0.826 & 0.827 \\
7B+32B (medium) & 0.852 & 0.878 & 0.865 \\
7B+32B (high) & 0.861 & 0.891 & 0.876 \\
\midrule
\textbf{Average} & \textbf{0.840} & \textbf{0.825} & \textbf{0.832} \\
\bottomrule
\end{tabular}
\caption{Per-stage classification performance of the sufficiency classifier on MATH (5,000 test samples).}
\label{tab:classifier_metrics}
\end{table}

\textbf{Model Selection} (ensemble before inference) approaches select the most appropriate model before generation. For example, Fly-swat or Cannon (FORC) uses a DistilBERT model to predict which LLM performs best for each query, enabling cost-effective model selection that reduces inference costs while maintaining performance~\cite{forc}. RouterEval combines LLM routing with recommender systems by collecting correctness records from thousands of models across datasets, training a router that outperforms individual strong models when selecting among about 10 candidates~\cite{RouterEval}. While model selection can reduce computational costs by choosing smaller models when appropriate, training robust routers requires extensive labeled data. 
% Without sufficient samples, the ensemble performance rarely exceeds that of using a single strong model. Additionally, these approaches face generalization challenges in zero-shot scenarios where queries differ significantly from training distributions.

\textbf{Output Aggregation} (ensemble during inference) methods integrate outputs of multiple LLMs during generation to leverage the knowledge of all models. These methods can be further divided into three types. Firstly, token-level methods ensemble multiple LLMs by aligning vocabularies and averages token distributions~\cite{gac,UniTe}. Secondly, sentence-level approaches choose the best piece from multiple candidate text segments based on quality metrics or voting mechanisms~\cite{SweetSpan}. Thirdly, response-level methods generate complete responses from each model and then select the final output with ranking models, or use a separate fusion model to combine insights from all responses~\cite{LLM-Blender}. Although these ensemble methods can improve response quality, they incur computational overhead as all LLMs perform inference for each query. This issue becomes more severe with the emergence of reasoning-intensive models like GPT4-o1~\cite{o1}, which employ extended chain-of-thought processes that can lead to overthinking problems~\cite{overthinking}.

\section{Non-thinking Mode Collaboration}
\label{app:non-thinking}

Table~\ref{tab:non-thinking} demonstrates that \mname also generalizes effectively to non-thinking LLMs. Without explicit reasoning chains, the 7B+32B collaboration achieves the highest average accuracy of \textbf{82.56\%}, surpassing both the 7B model at 80.40\% and the 32B model at 79.98\%, while reducing computational cost by \textbf{36.7\%} relative to the 32B baseline. On Precalculus, one of the more challenging subjects, accuracy reaches \textbf{73.81\%}, exceeding both single-model baselines by over 5 percentage points. These results confirm that our cost-aware collaboration framework operates independently of the reasoning paradigm, achieving superior accuracy-efficiency trade-offs through dynamic computation allocation.

\section{Prompt Schema Template for Thinking Insight Generation}
\label{app:prompt_schema}

\noindent Figure~\ref{fig:prompt_schema} summarizes the prompt schema used in our mentor--intern collaboration to elicit a compact \texttt{[Thinking Insights]} block.
The mentor LLM outputs exactly four structured components (Goal, Planning, Retrieval, Action), while the intern SLM produces the final answer based on these insights, enabling consistent downstream parsing and stable guidance.

\section{Self-competence Awareness of SLMs}
\label{app:classifier}

A key question in our framework is whether the SLM can reliably assess its own ability to solve a problem given the current guidance. We compare three approaches for predicting sufficiency, all using MLP classifiers with different inputs: (1) \textbf{Statistics}: perplexity and entropy features computed from the SLM's output probabilities; (2) \textbf{Hidden}: the last-layer hidden state at the final token position; and (3) \textbf{Hidden w/ LoRA}: same as Hidden, but additionally finetuning the SLM with LoRA before extracting hidden states.

As shown in Table~\ref{tab:classifier}, Statistics and Hidden methods achieve comparable performance, suggesting that the SLM possesses a degree of self-awareness regarding its own competence. Surprisingly, the lightweight Statistics approach, which only requires output probabilities, matches the performance of Hidden, which requires access to internal representations. This indicates that the SLM's confidence signals are already well-encoded in its output distributions.

Notably, Hidden w/ LoRA performs worse than the non-finetuned approaches, likely because finetuning on a limited dataset leads to overfitting. These findings validate our design choice of using perplexity and entropy as sufficiency indicators.

\section{Extended Analysis}
\label{app:analysis}

\subsection{Classifier Accuracy}
\label{app:classifier_acc}

Table~\ref{tab:classifier_metrics} reports the full performance of the sufficiency classifier evaluated on 5,000 MATH test samples. The classifier achieves an average F1 of 0.832 with precision consistently above 0.82 across all stages. The high precision indicates that when the classifier decides to stop early, it is correct 84\% of the time.

\subsection{Unified vs.\ Per-Subject Classifier}
\label{app:unified}

We investigate whether a single unified classifier trained on all MATH subjects can replace per-subject classifiers. As shown in Table~\ref{tab:unified}, the unified classifier slightly outperforms per-subject classifiers on average (+0.46\%), improving on 6 out of 7 subjects. This confirms that a single classifier is sufficient for deployment, eliminating per-subject training overhead.

\begin{table}[ht]
\centering
\resizebox{\linewidth}{!}{
\begin{tabular}{@{}lccc@{}}
\toprule
\textbf{Subject} & \textbf{Per-Subject} & \textbf{Unified} & \textbf{Gap} \\ \midrule
Algebra & 96.04 & 96.55 & +0.51 \\
Counting \& Prob. & 82.70 & 84.39 & +1.69 \\
Geometry & 73.07 & 72.44 & $-$0.63 \\
Intermediate Alg. & 67.77 & 69.43 & +1.66 \\
Number Theory & 88.70 & 89.44 & +0.74 \\
Prealgebra & 92.65 & 94.03 & +1.38 \\
Precalculus & 71.98 & 73.26 & +1.28 \\
\midrule
\textbf{Average} & \textbf{83.46} & \textbf{83.92} & \textbf{+0.46} \\
\bottomrule
\end{tabular}
}
\caption{Cascade accuracy (\%) comparison: per-subject vs.\ unified classifier.}
\label{tab:unified}
\end{table}

\subsection{Feature Importance}
\label{app:feature}

Table~\ref{tab:feature_importance} reports the top-10 feature importance scores from the GradientBoosting classifier. The importance is distributed relatively evenly (0.05--0.08), with entropy features (6 of top 10) and log-probability features (4 of top 10) contributing comparably, confirming that both signal families are informative.

\begin{table}[ht]
\centering
\begin{tabular}{@{}clc@{}}
\toprule
\textbf{Rank} & \textbf{Feature} & \textbf{Importance} \\ \midrule
1 & entropy\_trend\_change & 0.084 \\
2 & entropy\_min & 0.075 \\
3 & entropy\_last\_quarter & 0.069 \\
4 & log\_prob\_std & 0.066 \\
5 & log\_prob\_slope & 0.063 \\
6 & entropy\_std & 0.059 \\
7 & entropy\_mean & 0.057 \\
8 & log\_prob\_increase\_ratio & 0.057 \\
9 & log\_prob\_last\_quarter & 0.054 \\
10 & log\_prob\_trend\_change & 0.053 \\
\bottomrule
\end{tabular}
\caption{Top-10 feature importance scores from the sufficiency classifier.}
\label{tab:feature_importance}
\end{table}

\subsection{Stage Distribution}
\label{app:stage}

Table~\ref{tab:stage_distribution} shows how the cascade classifier distributes samples across stages for each MATH subject. Easier subjects like Prealgebra allocate 75.6\% of samples to the medium stage with only 6.5\% requiring high guidance, while harder subjects like Intermediate Algebra and Number Theory route over 50\% of samples to the high stage. This confirms that the cascade adaptively allocates reasoning effort based on problem difficulty.

\begin{table}[ht]
\centering
\resizebox{\linewidth}{!}{
\begin{tabular}{@{}lcccc@{}}
\toprule
\textbf{Subject} & \textbf{SLM} & \textbf{Low} & \textbf{Med} & \textbf{High} \\ \midrule
Algebra & 5.81 & 20.22 & 32.60 & 41.36 \\
Counting \& Prob. & 8.23 & 16.03 & 27.64 & 48.10 \\
Geometry & 8.77 & 19.83 & 38.41 & 32.99 \\
Intermediate Alg. & 12.51 & 8.64 & 26.02 & 52.82 \\
Number Theory & 2.41 & 1.30 & 36.67 & 59.63 \\
Prealgebra & 12.51 & 5.40 & 75.55 & 6.54 \\
Precalculus & 23.99 & 13.37 & 29.12 & 33.52 \\
\midrule
\textbf{Overall} & \textbf{9.38} & \textbf{11.52} & \textbf{31.54} & \textbf{47.56} \\
\bottomrule
\end{tabular}
}
\caption{Stage distribution (\%) across MATH subjects. The cascade classifier adaptively routes samples to different guidance levels.}
\label{tab:stage_distribution}
\end{table}

\subsection{Latency Measurements}
\label{app:latency}

We measure wall-clock latency on 100 MATH samples using DeepSeek-R1-Distill-Qwen-32B and -7B served via vLLM. As shown in Table~\ref{tab:latency}, \mname achieves 1.8$\times$ speedup over Mentor Only (12.64s vs.\ 22.67s per sample) while achieving higher accuracy. The continual judgment overhead is negligible, adding only 0.02--0.25 seconds per stage ($<$2\% of total generation time).

\begin{table}[ht]
\centering
\begin{tabular}{@{}lcc@{}}
\toprule
\textbf{Method} & \textbf{Avg Latency (s)} & \textbf{Speedup} \\ \midrule
Intern Only (7B) & 8.62 & 2.6$\times$ \\
Mentor Only (32B) & 22.67 & 1.0$\times$ \\
\mname (cascade) & 12.64 & 1.8$\times$ \\
\bottomrule
\end{tabular}
\caption{Wall-clock latency comparison on MATH.}
\label{tab:latency}
\end{table}

\subsection{Failure Case Analysis}
\label{app:failure}

We analyze the cascade routing outcomes on 5,000 MATH test samples (Table~\ref{tab:failure}). Among the 827 incorrect samples, 93.1\% are due to premature stops (460) or late stops (310), both reducible with a better classifier. Only 57 samples (1.14\%) are fundamentally unsolvable at all stages, yielding an oracle accuracy of 98.86\%.

\begin{table}[ht]
\centering
\begin{tabular}{@{}lcc@{}}
\toprule
\textbf{Outcome} & \textbf{Count} & \textbf{\%} \\ \midrule
Correct routing & 4,173 & 83.46 \\
Premature stop & 460 & 9.20 \\
Late stop & 310 & 6.20 \\
Unsolvable & 57 & 1.14 \\
\bottomrule
\end{tabular}
\caption{Cascade routing analysis on 5,000 MATH test samples.}
\label{tab:failure}
\end{table}

\end{document}